%% file: 2024_main_v1.tex
\documentclass[sigconf,authorversion,nonacm]{acmart} 

\usepackage{multirow}
\usepackage{graphicx} 
\usepackage{url}

\usepackage{hyperref}  
\usepackage{natbib}  
\usepackage[inline]{enumitem}
\usepackage{algorithm}
\usepackage{algorithmic}
\usepackage{adjustbox}

\AtBeginDocument{%
  }

\begin{document}

\title{A Multimodal Social Agent}

\author{Athina Bikaki and Ioannis A. Kakadiaris}

\renewcommand{\shortauthors}{A Bikaki and IA Kakadiaris}

\input{0.Abstract,v1}
\balance


\maketitle

\input{1.Introduction,v1}

\input{2.RelatedWork,v1}
\input{3.Methods,v1}

\input{4.Results,v1}
\input{5.Discussion,v1}
\input{6.Conclusion,v1}

\bibliographystyle{ACM-Reference-Format}
\bibliography{2024_main_v1}

\end{document}

%% file: 0.Abstract,v1.tex
\begin{abstract}
In recent years, large language models (LLMs) have demonstrated remarkable progress in common-sense reasoning tasks. This ability is fundamental to understanding social dynamics, interactions, and communication. However, the potential of integrating computers with these social capabilities is still relatively unexplored. This paper introduces MuSA, a multimodal LLM-based agent that analyzes text-rich social content tailored to address selected human-centric content analysis tasks, such as question answering, visual question answering, title generation, and categorization. It uses planning, reasoning, acting, optimizing, criticizing, and refining strategies to complete a task. Our approach demonstrates that MuSA can automate and improve social content analysis, helping decision-making processes across various applications. We have evaluated our agent’s capabilities in question answering (using the dataset HotpotQA), title generation (using the dataset WikiWeb2M), and content categorization (using the dataset MN-DS) tasks. MuSA performs substantially better than our baselines.
\end{abstract}

%% file: 1.Introduction,v1.tex
\section{Introduction}

The concept of autonomous LLM-based agents is already well established, and several successful examples exist in various domains. LLMs have shown great promise in acting as general-purpose decision-making machines, thus called \textit{language agents}, i.e., systems that use LLMs as a core computation unit to reason, plan, and act. To be successful in decision-making processes, LLM-based agents must be able to accomplish a given task autonomously. Although LLMs excel at retrieving information, they have limited reasoning skills, often finding it challenging to make logical inferences about social situations and comprehend social contexts and interactions. A key attribute in improving their responses through successive attempts is the application of advanced reasoning techniques.

Formulating high-quality prompts is foundational for agent interactions, impacts their performance, and poses a challenge for non-AI experts. Existing research on prompt optimization techniques suggests mainly optimization principles based on deep reinforcement learning. One disadvantage of deep reinforcement learning is the large number of samples required to learn an effective policy, which can be impractical in complex data collection scenarios. Prompting techniques and systematic reasoning advancements have recently improved agents' search, information retrieval, and reasoning capabilities. Verbal reinforcement, precisely the Reflexion \citep{shinn_reflexion_2023} method, helps agents learn from prior failings and improves learning over previous methods based on traditional reinforcement learning with gradient descent optimization. To facilitate the intercommunication of multi-agent systems and avoid hard-coding prompt templates in more complex systems, such as pipelines, new programming models have emerged that can abstract LLM calls and be parameterized to learn how to apply prompting, fine-tuning, and reasoning. State-of-the-art frameworks, such as DSpy \citep{khattab_dspy_2024} and TextGrad \citep{yuksekgonul_textgrad_2024}, have shown promising results in prompt optimization and overall system performance. The results have shown that they can substantially improve zero-shot accuracy, thus helping LLM agents answer complex questions efficiently and elicit sophisticated behavior.

\begin{figure}[!htbp]
  \centering
  \includegraphics[width=0.46\textwidth]{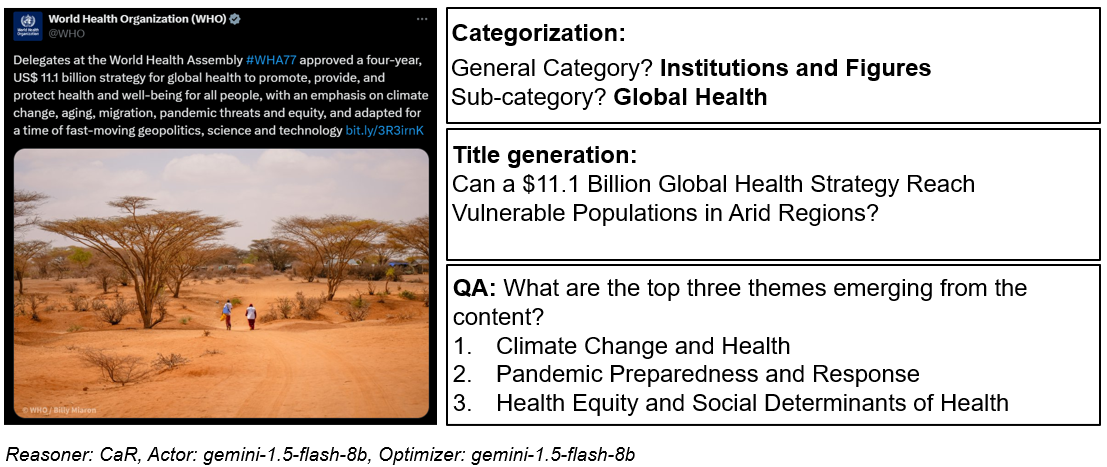}
   \vspace{-5pt}
   \caption{An example for MuSA. Post was selected from X \citep{world_health_organization_who_who_delegates_2024}.}\label{fig:headline-MuSA}
   \Description{An example for MuSA. Post was selected from X \citep{world_health_organization_who_who_delegates_2024}.}    
\end{figure}

While prompt quality is crucial, LLM's scope is further limited by the recency, validity, and breadth of their training data  \citep{qiu_llm-based_2024}. Their limited context window and lack of persistent memory lead to difficulties maintaining continuity across more extended interactions or agent communication \citep{qiu_llm-based_2024}. Furthermore, LLMs have pre-existing knowledge of the world, limited reasoning skills, and cannot usually handle other modalities than text. These limitations restrict their applications in the social science domain, which often requires reasoning skills beyond linguistic proficiency and information retrieval across modalities. In contrast, LLM-based agents can adapt dynamically based on their programmed knowledge and typically possess adaptive memory systems that facilitate their communication capabilities.

The contribution of our work is MuSA\footnote{Inspired from Latin Musa, Greek Mousa, Old French Muse, i.e., inspiring goddesses of the various arts such as music, dance, poetry, literature, and science (Greek Mythology).}, a multimodal LLM-based agent to analyze online social content. Social content analysis requires critical thinking and reasoning skills to understand different viewpoints and perspectives within a social context. We have equipped MuSA with these skills. While several multimodal LLM-based agents exist, to our knowledge, MuSA is the first to specialize in social content analysis. Its benefits can be summarized as follows:
\begin{enumerate}
  \item MuSA is designed to be modular and extensible. Currently, it supports five independent LLM-based units: reason, plan, optimize, criticize, refine, and act, which can be combined in various ways to meet different task requirements.
  \item MuSA is model-agnostic and can incorporate various models to support different capabilities or cost requirements.
  \item MuSA is optimized for a pre-selected number of social content analysis tasks (i.e., question answering, visual question answering, title generation, and content categorization), exhibits multimodal capabilities, and leverages complex internal processes to reason, such as chain-of-thought (CoT) \citep{wei_chain--thought_2022,kojima_large_2024} and self-reflection \citep{shinn_reflexion_2023}.
  \item It uses prompt and plan optimization through TextGrad  \citep{yuksekgonul_textgrad_2024} in all interactions to improve its performance in social content analysis. 
\end{enumerate}

The rest of the paper is organized as follows: Section 2 discusses related work. Section 3 explains the methods, Section 4 describes the results obtained, Section 5 discusses the findings, and Section 6 concludes this article.

%% file: 2.RelatedWork,v1.tex
\section{Related Work}

\noindent\textbf{Reason.} Advances in prompting methods, such as CoT \citep{wei_chain--thought_2022,kojima_large_2024}, and ReACT \citep{yao_react_2023}  have demonstrated capabilities to carry out several steps of reasoning traces to derive answers through proper prompting (Reason-only or Reason+Act).
Furthermore, recent in-context LLM agent-based learning techniques, such as Reflexion \citep{shinn_reflexion_2023} and Self-Refine \citep{madaan_self-refine_2023} (allows for iterative self-reflection to correct previous mistakes), have demonstrated the feasibility of autonomous decision-making agents built on top of an LLM. These methods use in-context examples or verbal reinforcement to teach agents to learn. Specifically, Reflexion converts binary or scalar feedback from the environment into verbal feedback in the form of a textual summary, which is then added as an additional context for the LLM agent in the subsequent trial. These methods significantly improve LLM's ability to learn in context and improve its performance, robustness, and reliability for the tasks at hand. The advancements in complex linguistic reasoning tasks have triggered an interest in complex visual reasoning tasks. Multimodal LLMs (MLLMs) have succeeded in text-rich content understanding tasks, such as image captioning, visual question answering, and optical character recognition. Recently, successful reasoning techniques such as in-context learning and chain-of-thought have been applied to the visual domain \citep{lan_improving_2023,shao_visual_2024} to improve the reasoning capabilities of MLLMs. Therefore, we have also designed MuSA's computing units to support multimodal models for specific actions to extend its cognitive abilities.

\noindent\textbf{Plan.} Planning is one of the most critical capabilities of agents and requires complicated understanding, reasoning, and decision-making progress \citep{nau_automated_2004}. Conventional methods rely on (deep) reinforcement learning or policy learning. However, these methods have several limitations, as they require expert knowledge, have limited fault tolerance, and require many examples to train and optimize, making them impractical \citep{huang_understanding_2024}. Recently, with the emergence of LLMs, integrating them as the cognitive core of agents has shown promising results in improving their planning ability. Specifically, techniques such as CoT \citep{wei_chain--thought_2022,kojima_large_2024} and ReACT \citep{yao_react_2023} are used more frequently in step-level planning (task decomposition) and execution. Other approaches, such as Reflexion \citep{shinn_reflexion_2023} and Self-Refine \citep{madaan_self-refine_2023}, are mainly used to improve the agent's planning ability. Lastly, methods such as REMEMBERER~\cite{zhang_rememberer_2023} and MemoryBank~\cite{zhong_memorybank_2024} incorporate an additional memory module, in which information is stored and retrieved during planning. Several additional methods are available that involve selecting the best plan from a pool of plans or using external planners, which are outside the scope of this work. The concepts mentioned can be combined, and in our study, we have employed various techniques to develop MuSA.

\noindent\textbf{Optimize.} LLMs are known to be sensitive to how they are prompted for each task, and optimization is usually done manually by trial and error, resulting in hand-crafted task-specific prompts that often require a deeper understanding of a particular domain and are challenging to maintain for non-AI experts. Recent efforts to automate prompt tuning are encouraging, and various approaches are being explored to achieve this goal. For instance, the Optimization by PROmpting (OPRO) \citep{yang_large_2024} approach treats LLMs as optimizers. The EVOPrompt \citep {guo_connecting_2024}, a discrete prompt optimization framework, uses LLMs as evolutionary operators and is based on evolutionary algorithms to guide the optimization process. However, these methods focus solely on learning prompts and may require additional training data to learn. On the contrary, generic frameworks, such as DSpy \citep{khattab_dspy_2024} and TextGrad \citep{yuksekgonul_textgrad_2024} provide a more systematic approach for developing and optimizing multi-stage LM-based pipelines and agents that are domain, model, and input agnostic. Our MuSA uses prompt and plan optimization through TextGrad \citep{yuksekgonul_textgrad_2024} in all interactions to maintain excellent performance in social content analysis.

\noindent\textbf{Criticize.} LLMs occasionally may exhibit hallucinations, faulty code, or even toxic content, and to address these challenges, the CRITIC \citep{guo_connecting_2024} method is introduced that uses LLMs to verify and rectify their output through external tools. Our work introduces a role akin to generating natural language critiques of plan outputs. Our critic functions as a form of assistance in preventing infeasible plans.

\noindent\textbf{Refine.}
LLMs do not always generate the best output on their first attempt. Self-Refine \citep{madaan_self-refine_2023} is an approach that allows the LLM to improve the initial output through iterative feedback and refinement. Self-Refine uses the same LLM to generate, get feedback, and refine its outputs. In our work, we define the refiner as responsible only for refinement from feedback from the critic, and the planner handles the initial generation. This provides greater flexibility in orchestrating models with different capabilities.

\noindent\textbf{Act.} LLMs often over-simplify or misrepresent textual or visual nuances, leading to a less accurate understanding of the content. Moreover, in many cases, LLMs struggle to produce the correct answer consistently in one shot. The application of Retrieval-Augmented Generation (RAG) technology has emerged to address these limitations, allowing for more precise responses \citep{lewis_retrieval-augmented_2020}. Approaches that repurpose knowledge from pre-trained models using few-shot prompting can enhance their responses. LLMs have been shown to outperform previous zero-shot methods and can sometimes become competitive with prior state-of-the-art fine-tuning approaches \citep{brown_language_2020,yang_harnessing_2024}. However, these methods cannot effectively handle complex tasks that require the model to self-improve by learning from its own mistakes. 

\begin{figure*}[!htbp]
    \begin{center}
   \includegraphics[width=\textwidth]{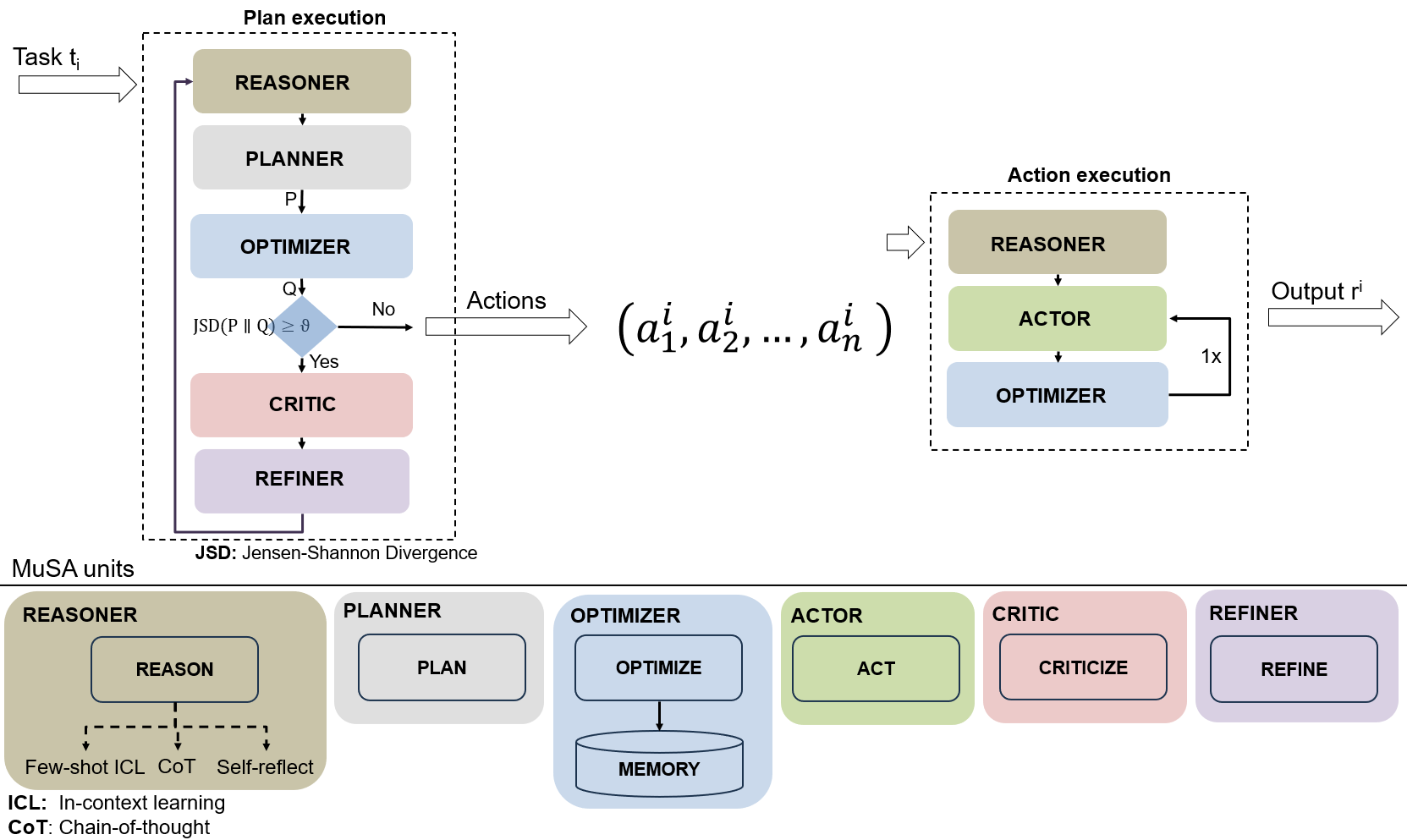}
   \caption{Visualization of MuSA plan and action execution for our selected content analysis tasks (T). The similarity between the responses of the planner and optimizer is assessed using Jensen-Shannon divergence (JSD). MuSA available units (B).}\label{fig:agent-process}
    \Description{Visualization of MuSA plan and action execution for our selected content analysis tasks (T). The similarity between the responses of the planner and optimizer is assessed using Jensen-Shannon divergence (JSD). MuSA available units (B).}
    \end{center}
  \end{figure*}

%% file: 3.Methods,v1.tex
\section{Methods}
This section outlines the architecture of MuSA~(\autoref{fig:agent-process}). First, we describe at a high level the coordinating units necessary to complete a task. Next, we elaborate on the functionalities of each unit that are part of MuSA's computing engine.

For our work, we assume a closed static environment $E$. We define a \textit{task} $t$ as a set of literals that describe a goal necessary for the agent(s) to build a plan. A plan $\pi=(a_1,a_2,...,a_n)$ where $a_i\in A$ for each $i$, and $A$ is the action space, can involve a single action or a sequence of actions coordinated by agent(s). Given a task $t_i$, the task adaptation and solving procedure by a single MuSA agent is shown in Algorithm~\ref{algo:algorithmMuSA}.

\subsection{Reasoner} \label{sec:irm}
MuSA has been equipped with in-context few-shot learning capabilities, CoT, and self-reflective behavior to improve reasoning and decision-making. These prompting techniques can be activated since not every reasoning method suits every task. Few-shot and zero-shot CoT prompting allow LLMs to generate intermediate reasoning steps explicitly before predicting the final answer. In the few-shot scenario, this is done by providing a few demonstration examples, while in the zero-shot scenario, it is accomplished without any demonstration examples. In the latter case, this is achieved by appending the phrase "let's think step by step" to the input. Self-reflection was applied to the input with a single strategy: "apply reflection to the following reasoning trace." The CoT and/or self-reflected responses are then used as input to the optimization process.

\subsection{Planner} Current LLM agent planning solutions aim to map tasks to sequences of executable actions. Similarly, the planner is responsible for constructing a plan that consists of a sequence of executable social content analysis actions. Closely related to our work \textit{Reflection and Refinement} \citep{shinn_reflexion_2023,madaan_self-refine_2023,gou_critic_2024,huang_understanding_2024} methodologies encourage LLM to reflect on failures and refine the plan. We have introduced four new processes in planning: (i) reason, (ii) optimize, (iii) criticize, and (iv) refine. The entire process is described in~\autoref{eq:MuSA-planning}:

\begin{equation}
    \label{eq:MuSA-planning}
    \begin{split}    
    &s_0= \pi(E, t, \Theta, p^{rsn})\\   
    &o_i = o(E, t, s_i, \Theta, p^{rsn})\\
    &c_i = c(E, t, s_i, o_i, \Theta, p^{rsn})\\
    &r_{f_i}=r_f(E, t, c_i, \, p^{rsn})\\
    &s_{i+1}=\pi(E, t, r_{f_i},\Theta, p^{rsn})\\
    \end{split}
\end{equation}

\noindent where $s_0$ is the initial state, $\pi$ is the plan process, $E$ is the environment, $t$ the task, $\Theta$ the set of LLM parameters, $p$ the prompt for the task, $rsn$ denotes the reasoning strategy attached to the prompt, $i = \{0, 1,...,n-1\}$ represents the time step, $o$ denotes the optimize process, $c$ denotes the criticize process, and $r_f$ denotes the refine process. The output is a set of actions and instructions (prompts) to be executed.

\begin{algorithm}
    \caption{MuSA task solving}\label{algo:algorithmMuSA}
    \begin{algorithmic}[1]
        \STATE \textbf{Input} t, reasoner, planner, optimizer, critic, refiner, actor, trials, $\theta$ (threshold)
        \STATE \textbf{Output} r (response) //task-specific output
        \STATE r $\leftarrow \emptyset$        
        \STATE p${_t}$ $\leftarrow$ CREATE-FROM($t$)   //prompt        
        \STATE //plan execution
        \FOR{$t_i=0,...,$trials}  
        \STATE p$_r \leftarrow$ REASON($p_t$, reasoner) 
        \STATE $r_p \leftarrow$ PLAN(p${_r}$, planner)    
        \STATE $r_o \leftarrow$ OPTIMIZE($r_p$, optimizer)  // optimizer feedback
        \STATE $A \leftarrow r_o$ //actions
        
        \IF{$t_i<$trials-1}     
        \STATE P $\leftarrow$ ENCODE($r_p$)
        \STATE Q $\leftarrow$ ENCODE($r_o$)  
        
        \IF{$JSD(P || Q) \geq \theta$ }
            \STATE p$_u \leftarrow$ CREATE-FROM($r_p$, $r_o$)
            \STATE $p_c \leftarrow$ CRITICIZE(p$_u$, critic)
            \STATE $p_t \leftarrow$ REFINE($p_c$, refiner)
        \ELSE            
            \STATE break
        \ENDIF   
        
        \ENDIF
        \ENDFOR
        
        \STATE //action execution
        \FOR{$a_i \in A$}
            \STATE p $\leftarrow$ CREATE-FROM($a_i$)
            \STATE p$_r \leftarrow$ REASON(p, reasoner)            
            \STATE $r_a \leftarrow$ ACT(p$_r$, actor)   // actor response            
            \STATE $r_o \leftarrow$ OPTIMIZE($r_a$, optimizer)  // optimizer feedback
            \STATE $r_i \leftarrow$ ACT($r_o$, actor)   // actor response            
            \STATE r $\leftarrow (r_i,  r)$
        \ENDFOR
        \STATE \textbf{return} r
    \end{algorithmic}
\end{algorithm}

\subsection{Optimizer} 
To optimize the response obtained from the planner or actor, we have employed TextGrad \citep{yuksekgonul_textgrad_2024}. This automatic differentiation framework works similarly to the backpropagation method for training neural networks. Still, it operates using only textual information (i.e., prompts) and uses LLM API calls or other external callers to propagate gradients. In that case, the gradients represent feedback on improving the input. This process involves performing a forward pass with a selected engine (e.g., LLM API call, web search), computing the TextLoss \citep{yuksekgonul_textgrad_2024}, collecting the gradients, and then updating the answer using the Textual Gradient Descent (TGD) optimizer (Equations \ref{eq:TextGrad gradients} and~\ref{eq:TextGrad updated prompt}) \citep{yuksekgonul_textgrad_2024}. The TextLoss in our examples is defined as "critical evaluation instructions and analysis of the reflected input along with the initial questions."

 \begin{equation}
    \label{eq:TextGrad gradients}
    \begin{split}
    \frac{\partial \mathcal{L}}{\partial x^{rsn}} = \nabla_{LLM}(& x^{rsn}, y, \frac{\partial \mathcal{L}}{\partial y} ) 
    \end{split}
    \end{equation}
    
    \begin{equation}
        \label{eq:TextGrad updated prompt}
        x_u = x^{rsn} - \alpha\frac{\partial \mathcal{L}}{\partial x^{rsn}} 
 \end{equation}

\noindent where the symbol $x^{rsn}$ is the variable we want to optimize (e.g., prompt, plan) with the reasoning strategy included, $y$ the prediction, $\mathcal{L}$ the objective function, and $\alpha$ the learning rate. All inputs are text.

The reasoning mechanisms (e.g., CoT, self-reflection) are applied to the actor or planner before invoking the optimizer. In that way, we propagate reasoning, for example, the self-reflect mechanism in the optimization process without explicitly calling it. The optimization can be set to run for a specified number of iterations. In reasoning mechanisms that benefit from repetitions, such as self-reflection, the number of repetitions can align with the number of iterations in TextGrad. For this process to work optimally, the optimizer must be capable of self-reflecting and improving.

\subsection{Critic}

A critic model is employed when the dissimilarity between the planner's and optimizer's responses exceeds a specified threshold $\theta$. The critic evaluates the optimized response obtained through TextGrad \citep{yuksekgonul_textgrad_2024} against the planner's response and selects the most suitable response. The activation of the critique mechanism relies on a task-related decision rubric and is controlled by an input threshold $\theta$. To achieve this, the embeddings of both responses are calculated, and their similarity is assessed using Jensen-Shannon divergence (JSD) \citep{lin_divergence_1991}. In cases where the similarity of responses falls below the threshold $\theta$, a critic will not be called.

\subsection{Refiner} 

The refining process gets feedback from the critic to refine the plan before execution. When the critic provides actionable feedback, the refiner translates the input into actionable instructions for the planner based on the feedback. Then, the entire process cycle starts again. Currently, the system allows only one trial before exiting and running with the existing plan, as the planner and optimizer may not converge.

\subsection{Actor}
We define an action space $A=\{a_1,...,a_n\}$ for social content analysis. By combining different actions, MuSA can accomplish various tasks. MuSA receives information about a selected action through prompts. \autoref{table-MuSA-actions} presents the supported actions. 

\begin{table}[H] 
    \caption{Action space for MuSA to interact through. MM denotes multimodality, and ET denotes the use of external tools.}
     \vspace{-10pt}
    \label{table-MuSA-actions}
    \resizebox{\columnwidth}{!}
    {\begin{tabular}{|r|l|c|c|p{3.5in}|} 
    \hline
   \textbf{Id} & \textbf{Action} & \textbf{MM} & \textbf{ET} &\textbf{Description} \\
    \hline    
    1&Question answering (QA) & \checkmark & $\times$ & Answers natural language questions about textual content\\    
    2&Visual Question answering (VQA) & \checkmark & $\times$ & Answers natural language questions about text-rich content\\    
    3&Title generation & \checkmark & $\times$ & Creates a short headline from the input\\
    4&Categorization & \checkmark & $\times$ & Classifies and organizes online content into categories. It can also be used for sub-categories\\
    \hline    
    \end{tabular}}
\end{table}

\subsection{External Tools}
In specific actions, external knowledge is necessary, and we provide this specialized knowledge through external tools (\autoref{table-MuSA-actions}). This functions as a static read-only memory that cannot be updated and is more akin to semantic memory. Semantic memory is responsible for storing factual information about the world and plays a vital role in memory \citep{lindes_toward_2016}. We aim to enhance MuSA's reasoning capabilities by providing this specialized external knowledge.

%% file: 4.Results,v1.tex
\section{Results}

\begin{figure}[H]
  \centering
    \adjustbox{valign=c}{\includegraphics[width=0.46\textwidth]{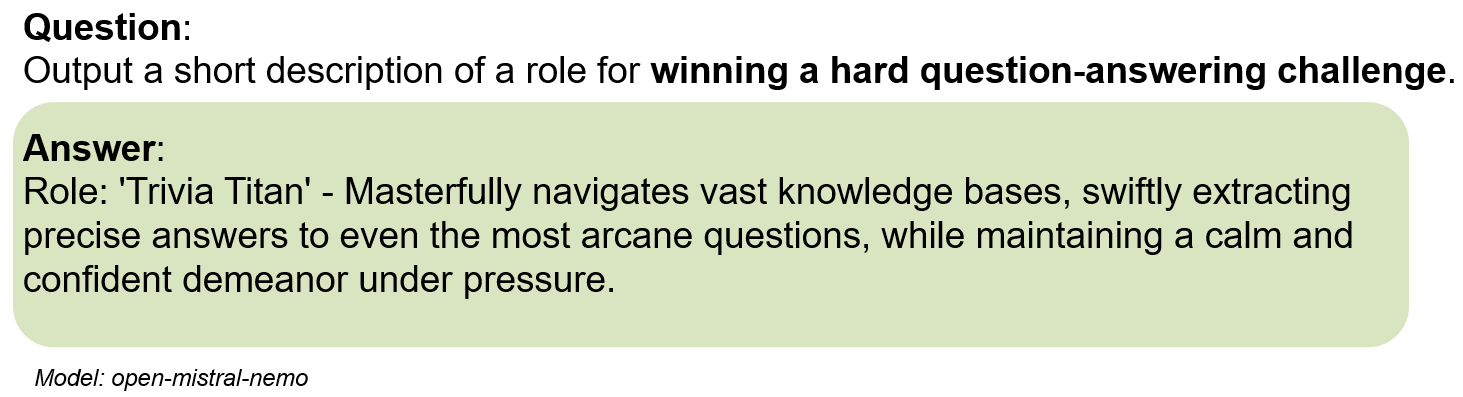}}
  \vspace{-5pt}
  \caption{Role assignment for a task.}\label{fig:prompt_role_descr}
  \Description{Role assignment for a task.}
\end{figure}

\noindent\textbf{Environment setup.} Initially, we assign a (different) LLM to the actor, optimizer, critic, and refiner units. Then, we define a task $t$ and ask MuSA to generate an appropriate role description for the task at hand, which is used as a system role in all subsequent calls related to the task. For the role creation, we use a separate LLM that is not used as an actor, optimizer, critic, or refiner to avoid preference biases. An example is shown in~\autoref{fig:prompt_role_descr}.

\subsection{Experiments}
To test MuSA, we have done experiments for each action in~\autoref{table-MuSA-actions}.
In our experiments, we use the following acronyms: Re, A, and O denote reasoner, actor, and optimizer, respectively. CaR denotes zero-shot CoT and self-reflection reasoning strategy. 
For all our experiments, we set the temperature to $0$ for Gemini \citep{google_deepmind_gemini_2024} and $0.7$ for Mistral \citep{mistral_mistral_2024} models. We also set the nucleus sampling to $0.99$.

\subsubsection{Plan.}
\begin{figure}[H]
    \begin{center}
   \includegraphics[width=0.46\textwidth]{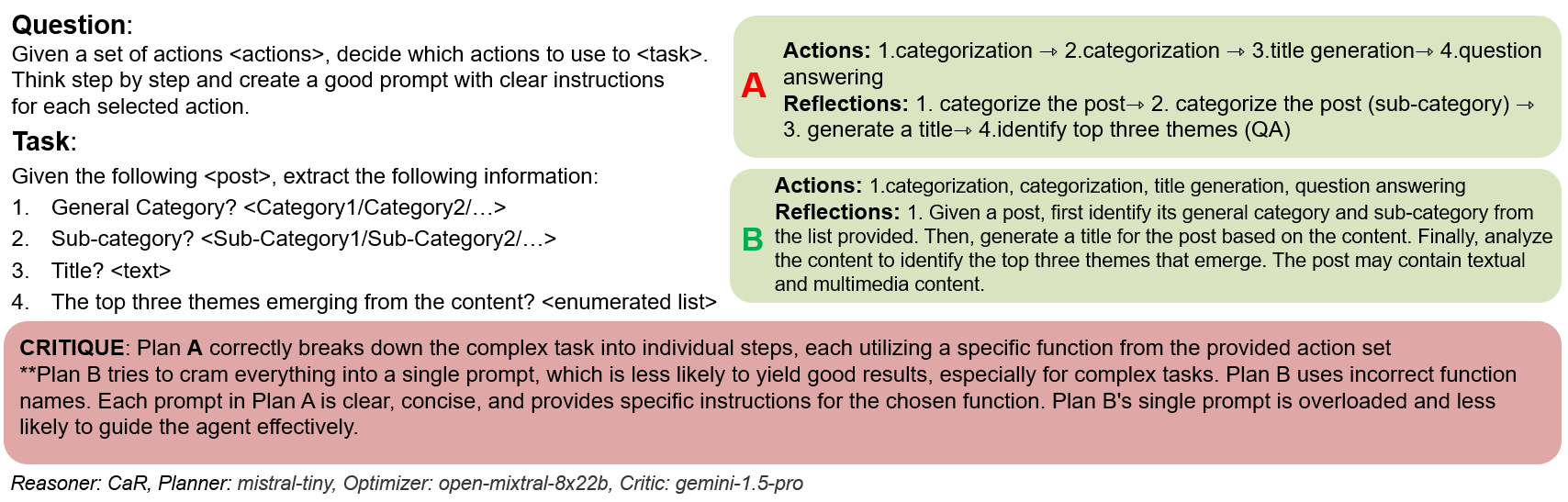}
    \vspace{-5pt}
    \caption{Decision plan of a task. Plan A was proposed by the planner, and plan B by the optimizer.}\label{fig:prompt_decision_plan}
    \Description{Decision plan of a task. Plan A was proposed by the planner, and plan B by the optimizer.}
    \end{center}
  \end{figure}

Planning is one of the most critical abilities of agents, and that often requires complicated understanding and reasoning. We provide real-world examples in Figures~\ref{fig:headline-MuSA} and \ref{fig:visual-example-FB} that show a composite task assigned to MuSA. The planner decides which actions to use to complete the task. The decision plan proposed by MuSA is shown in~\autoref{fig:prompt_decision_plan}. Initially, the planner proposes a plan (plan A), and the optimizer tries to optimize it (plan B). Since plans A and B have substantial differences, a critic model is introduced to criticize the proposed plans and provide feedback for improvement. In this example, the critic suggested using plan A, which better addresses the task. Since there is actionable feedback, the refiner creates refined instructions for the planner, and the entire process cycle starts again.

\subsubsection{Question answering (QA)} 
\begin{figure}[H]
    \begin{center}
   \includegraphics[width=0.46\textwidth]{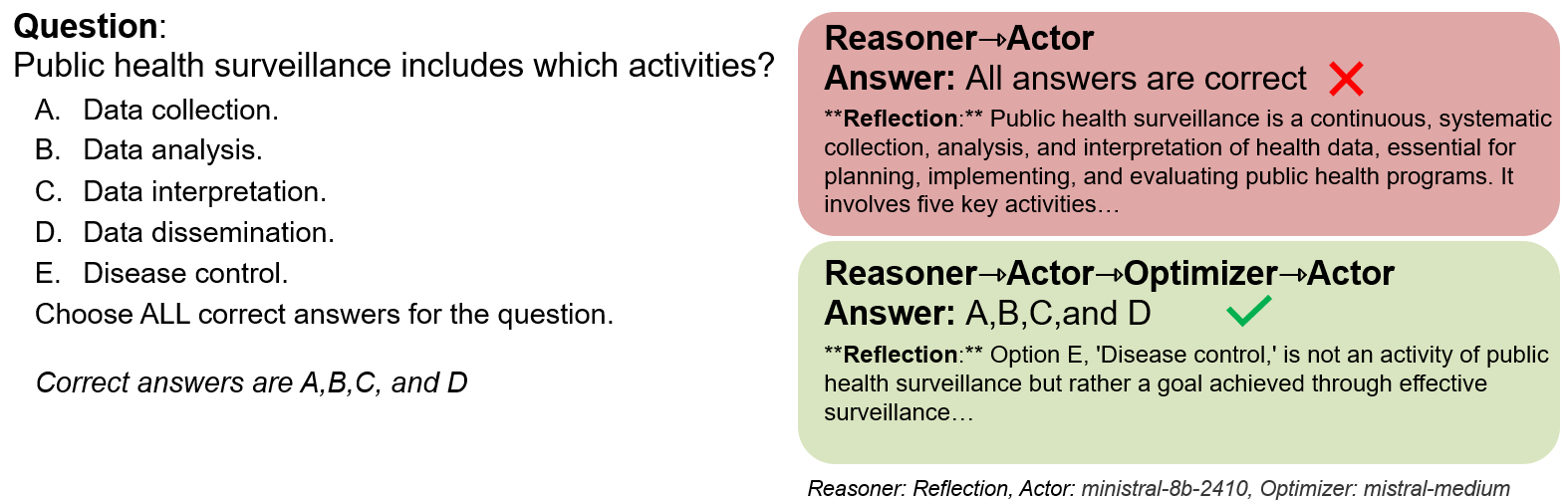}
    \vspace{-5pt}
    \caption{An example of a multiple-choice question from a quiz related to public health surveillance \citep{cdc_principles_2023}. A single trial with different roles within the same environment and task.}\label{fig:prompt_rflx_textgrad}
    \Description{An example of a multiple-choice question from a quiz related to public health surveillance \citep{cdc_principles_2023}. A single trial with different roles within the same environment and task.}
    \end{center}
  \end{figure}

The task of open-domain question answering, which corresponds to action 1 in~\autoref{table-MuSA-actions}, provides a measurable and objective way to test the reasoning ability of intelligent agents \citep{yang_hotpotqa_2018}. This task involves finding the answer to a question from an extensive text collection. Typically, it requires identifying a small set of relevant documents and extracting the answer from them. To illustrate how the actor and optimizer processes function in that case, we present an example in \autoref{fig:prompt_rflx_textgrad}, which contains only textual information but requires critical thinking and reasoning to answer the question. The optimizer improved the actor's self-reflected output and selected the correct answer.

Furthermore, we tested MuSA on the  HotpotQA \citep{yang_hotpotqa_2018} dataset to evaluate multi-hop reasoning. HotpotQA contains 113k Wikipedia-based question-answer pairs. In line with previous work \citep{shinn_reflexion_2023}, we randomly selected 100 hard questions from the training set for our evaluation that are closely related to the social domain, using the context provided. We have used the same evaluation metrics proposed by the authors \citep{yang_hotpotqa_2018}. Our results are presented in~\autoref{table-MuSA-results-QA}.

\begin{table}[H] 
\caption{Pass@1 100 on selected evaluation metrics HotpotQA \citep{yang_hotpotqa_2018}. EM, F1, P, R: denote Exact Match, F1-score, Precision, and Recall, respectively. We have used \text{Gemini-1.5-Flash-8B} \citep{google_deepmind_gemini_2024} for both the actor and optimizer.}
\vspace{-5pt}
\label{table-MuSA-results-QA}
{\begin{tabular}{l r r r r} 
\hline
\textbf{Method} & \textbf{EM} & \textbf{F1} & \textbf{P} & \textbf{R} \\
\hline    
Re$_{CaR}\rightarrow$A & 60.0 &\textbf{72.1} &\textbf{73.6} &73.2 \\
Re$_{CaR}\rightarrow$
A$\rightarrow$O$\rightarrow$A& \textbf{62.0} &67.6 &66.5& \textbf{78.5}\\ 
\hline
\end{tabular}}
\end{table}

\subsubsection{Visual question answering (VQA)}

\begin{figure}[H]
  \begin{center}
 \includegraphics[width=0.46\textwidth]{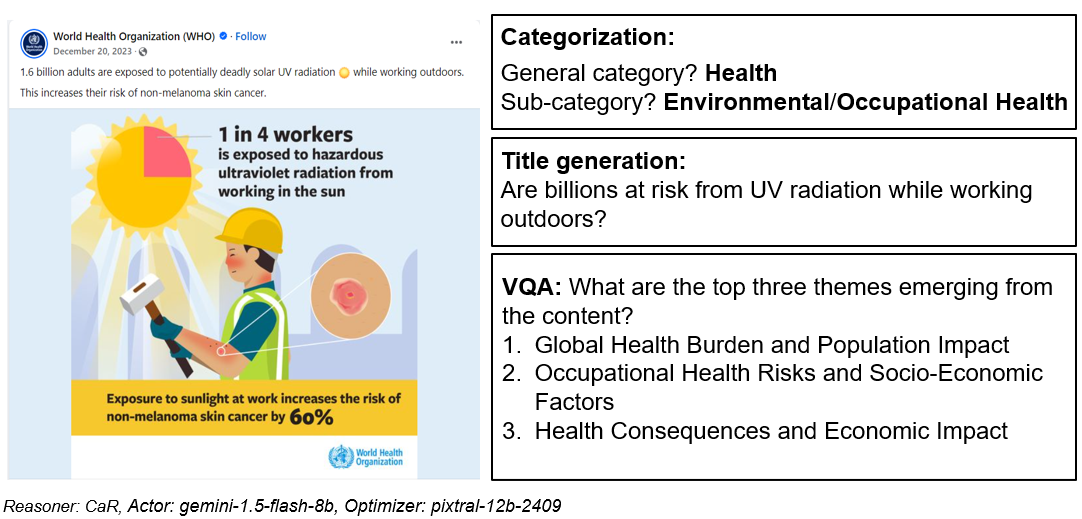}
  \vspace{-5pt}
  \caption{Example of a multimodal input where visual content has more information than text. Post was selected from Facebook \citep{world_health_organization_who_sun_2023}.}\label{fig:visual-example-FB}
  \Description{Example of a multimodal input where visual content has more information than text. Post was selected from Facebook \citep{world_health_organization_who_sun_2023}.}
  \end{center}
\end{figure}

The visual information in online content is as essential as textual information. MuSA is designed to process multimodal information in specific actions, as seen in~\autoref{table-MuSA-actions}. VQA task corresponds to action 2 in~\autoref{table-MuSA-actions}. MuSA may use a multimodal LLM for both the actor and optimizer.
The example in ~\autoref{fig:visual-example-FB} demonstrates how MuSA can help surface important information from images for publication. In this example, most of the information is contained in the image. Both actor and optimizer use multimodal LLMs.

\subsubsection{Title generation.}
\begin{figure}[H]
  \begin{center}
 \includegraphics[width=0.46\textwidth]{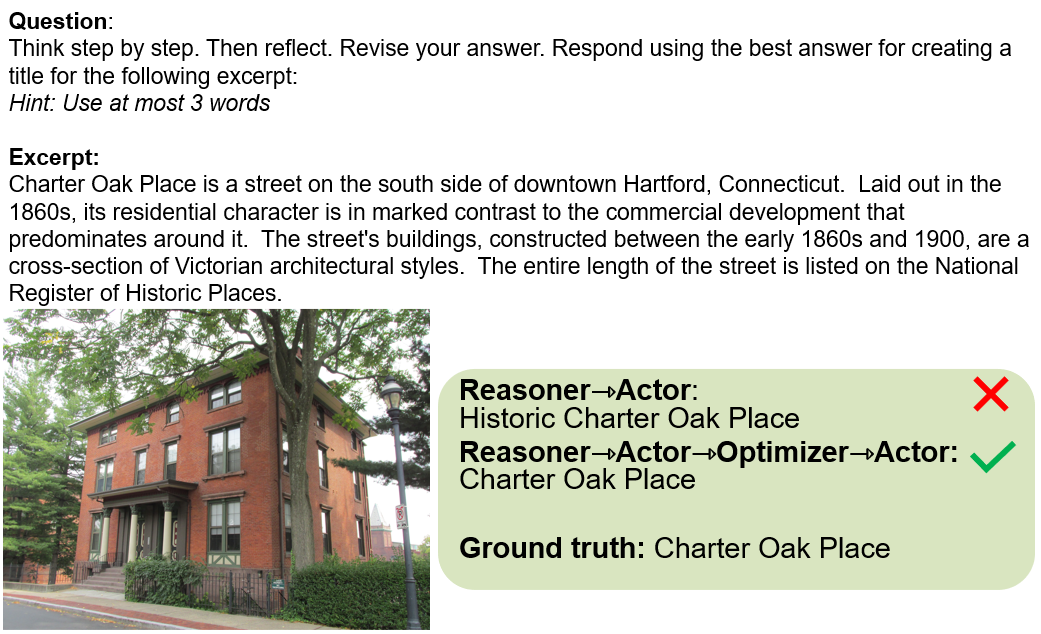}
  \vspace{-5pt}
  \caption{Title generation using an example from WikiWeb2M \citep{burns_suite_2023}.}\label{fig:webpage_understanding}
  \Description{Title generation using an example from WikiWeb2M \citep{burns_suite_2023}.}
  \end{center}
\end{figure}

Title generation is a common task in web content generation and analysis for creating concise and informative titles or headlines for content. To evaluate MuSA in the title generation task, which corresponds to action 3 in~\autoref{table-MuSA-actions}, we have used the Wikipedia Webpage suite (WikiWeb2M) \citep{burns_suite_2023} dataset, mainly introduced for studying multimodal webpage understanding. WikiWeb2M contains 2M pages with all the associated image, text, and structure data, but we randomly selected 100 web pages from the validation set for our study. MuSA actor is presented with the multimodal information from a section of a web page and is asked to generate a title (\autoref{fig:webpage_understanding}). 

The optimizer gets the actor's response and tries to optimize it using the available visual content. Finally, it provides suggestions and feedback for the actor to retry and generate a new response. Our evaluation results are presented in~\autoref{table-MuSA-results-WU}.

\begin{table}[H] 
  \caption{Pass@1 100 on selected evaluation metrics on WikiWeb2M \citep{burns_suite_2023}.  B4 and RL denote BLEU-4 and ROUGE-L scores, respectively. F1, P, R: denote F1-score, Precision, and Recall, respectively. The input is multimodal. For the actor, we have used the \text{Gemini-1.5-Flash-8B} \citep{google_deepmind_gemini_2024}, and for the optimizer, the Pixtral-12B-2409 \citep{mistral_mistral_2024} models.}
  \vspace{-5pt}
  \label{table-MuSA-results-WU}  
  \resizebox{\columnwidth}{!}
  {\begin{tabular}{l r r r r r} 
  \hline  
  \textbf{Method}&\textbf{EM} & \textbf{B4} & \textbf{RL F1} & \textbf{RL P}& \textbf{RL R}\\
  \hline    
  Re$_{CaR}\rightarrow$A& 26&39.9 &60.4 &58.6&68.2 \\
  Re$_{CaR}\rightarrow$A$\rightarrow$O$\rightarrow$A & \textbf{32}&\textbf{41.6}&\textbf{61.3}&\textbf{59.0}&\textbf{71.9} \\
  \hline
 \end{tabular}}
\end{table}

\subsubsection{Categorization}
\begin{figure*}[!htb]
  \begin{center}
 \includegraphics[width=\textwidth]{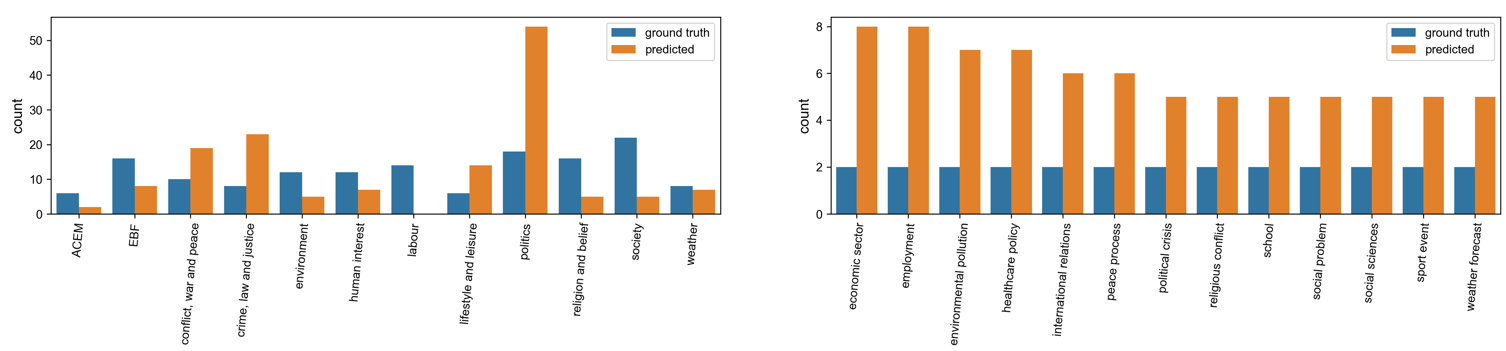}
  \vspace{-5pt}
 \caption{Most discrepant Level-1 (L) and Level-2 (R) categories from our selection of articles in the MN-DS \citep{petukhova_mn-ds_2023} dataset. ACEM stands for arts, culture, entertainment, and media, and EBF stands for economy, business, and finance.
  }\label{fig:CC_level_1_2_diffs}
  \Description{Most discrepant Level-1 (L) and Level-2 (R) categories from our selection of articles in the MN-DS \citep{petukhova_mn-ds_2023} dataset. ACEM stands for arts, culture, entertainment, and media, and EBF stands for economy, business, and finance.}
  \end{center}
\end{figure*}

Categorization involves classifying content into predefined categories. It corresponds to action \textit{Categorization} 4 in~\autoref{table-MuSA-actions}. We have used the MN-DS \citep{petukhova_mn-ds_2023} dataset, which contains 10,917 news articles with hierarchical predefined categories, to evaluate MuSA in content categorization. We have performed stratified sampling to create a subset of 218 articles from the test set, containing examples from 17 first-level and 109 second-level categories. First, we classified the data into first-level categories (Level-1 in~\autoref{table-MuSA-results-CC}). Then, we used these first-level categories to filter and classify into second-level categories (Level-2 in~\autoref{table-MuSA-results-CC}). In \autoref{fig:CC_level_1_2_diffs}, we show the first-level and second-level categories with the highest disagreements. For the second-level categories, we have selected to show the first 13 disagreements. A significant proportion of the articles were classified under the first-level category "politics," suggesting the need for further investigation.

\begin{table}[!ht] 
  \caption{Pass@1 218 Acc, F1, P, R on the MN-DS \citep{petukhova_mn-ds_2023} dataset. Acc, F1, P, R: denote Accuracy, F1-score, Precision, and Recall, respectively. For both the actor and optimizer we have used the \text{Gemini-1.5-Flash-8B} \citep{google_deepmind_gemini_2024} model.}  
  \vspace{-5pt}
  \label{table-MuSA-results-CC}
  \resizebox{\columnwidth}{!}
  {\begin{tabular}{l r r r r c r r r r} 
  \hline
  \textbf{Category} & \multicolumn{4}{c}{\textbf{Re$_{CaR}\rightarrow$A}} & &\multicolumn{4}{c}{\textbf{Re$_{CaR}\rightarrow$ A$\rightarrow$O$\rightarrow$A}}\\  
  \cline{2-5} \cline{7-10} 
  & \textbf{Acc} & \textbf{F1} &\textbf{P} & \textbf{R} & & \textbf{Acc} &\textbf{F1} & \textbf{P} & \textbf{R}\\
  \hline      
  Level-1 &48.1&43.5&48.2 &48.1 & &\textbf{50.8}&\textbf{47.1}&\textbf{53.6}&\textbf{50.8} \\
  Level-2 &54.5&48.3&49.6 &54.5 & &\textbf{56.4}&\textbf{50.7}&\textbf{52.7}&\textbf{56.4} \\
  \hline
 \end{tabular}}
\end{table}

\subsection{Ablation studies}

\subsubsection{QA: Impact of reasoning}
The actor's performance is evaluated with two internal reasoning mechanisms: (i) a standard LLM API call with zero-shot CoT instructions and (ii) a standard LLM API call with zero-shot CoT instructions and self-reflection. Zero-shot CoT appends the phrase "let's think step by step" to the prompt without demonstrating examples, and self-reflection uses a single attempt. The results are displayed in~\autoref{table-MuSA-ablation-results-reason}.  

\begin{table}[H] 
  \caption{Pass@1 100 EM, F1, P, R on HotpotQA \citep{yang_hotpotqa_2018}.  EM, F1, P, R: denote Exact Match, F1-score, Precision, and Recall, respectively.}
  \vspace{-5pt}
  \label{table-MuSA-ablation-results-reason}
  {\begin{tabular}{c l r r r r} 
  \hline 
  & \textbf{Method} &\textbf{EM} & \textbf{F1} & \textbf{P} & \textbf{R}\\
  \hline   
  \multirow{4}{*}{\rotatebox{90}{Open}} & Re$_{CoT}\rightarrow$A$_{\text{Mistral-tiny}}$ & 39.0 &\textbf{55.9} &55.4 &\textbf{66.9}\\
  & Re$_{CaR}\rightarrow$A$_{\text{Mistral-tiny}}$ & 38.0 &54.3 &54.2 &61.6\\ 
  & Re$_{CoT}\rightarrow$A$_{\text{Mistral-7B}}$ & \textbf{40.0} &55.7 &\textbf{55.7}&65.4\\
  & Re$_{CaR}\rightarrow$A$_{\text{Mistral-7B}}$ & \textbf{40.0} &55.4&55.5 &62.2\\
  \hline
  \multirow{4}{*}{\rotatebox{90}{Closed}} & Re$_{CoT}\rightarrow$A$_{\text{Gemini-Flash-8B}}$\footnotemark[1] &56.0 &71.4 &73.4 &72.5\\
  & Re$_{CaR}\rightarrow$A$_{\text{Gemini-Flash-8B}}$ &\textbf{60.0} &\textbf{72.1} &\textbf{73.6} &\textbf{73.2}\\
  & Re$_{CoT}\rightarrow$A$_{\text{Ministral-8B}}$ &59.0 &70.7& 72.9& 71.1\\
  & Re$_{CaR}\rightarrow$A$_{\text{Ministral-8B}}$ & 52.0 &60.5& 61.8& 61.7\\   
  \hline
  \end{tabular}}
\end{table}
\footnotetext[1]{Gemini version: 1.5}

The performance of different reasoning strategies seems to be model-specific. For example, Gemini-1.5-Flash-8B \citep{google_deepmind_gemini_2024} benefits greatly from CoT and self-reflection, while the Mistral family models \citep{mistral_mistral_2024} achieve the best performance when CoT only is used.

\subsubsection{QA: Impact of actor and optimizer roles} We evaluate the performance of the actor and optimizer roles using two settings: (i) the actor and optimizer use the same LLM, and (ii) the optimizer uses a stronger LLM. 
The selected actor's reasoning mechanism performs best as demonstrated in~\autoref{table-MuSA-ablation-results-reason}. 
We aim to improve the agent's reasoning performance while balancing performance, cost, and complexity. For our experiments, we have used as an actor smaller models up to 8B parameters and 8B or larger models for the optimizer. We also compare open and closed-source combinations of models for the actor and optimizer roles. The results are displayed in~\autoref{table-MuSA-ablation-results-actor-eval}.

\begin{table}[H] 
  \caption{Pass@1 100 EM, F1, P, R on HotpotQA \citep{yang_hotpotqa_2018}.  EM, F1, P, R: denote Exact Match, F1-score, Precision, and Recall, respectively.}
  \vspace{-5pt}
  \label{table-MuSA-ablation-results-actor-eval}
  \resizebox{\columnwidth}{!}
  {\begin{tabular}{c l r r r r} 
  \hline  
  & \textbf{Method}&\textbf{EM} & \textbf{F1} & \textbf{P} & \textbf{R}\\
  \hline   
  \multirow{4}{*}{\rotatebox{90}{Open}}   
  & Re$_{CoT}\rightarrow$ A$_{\text{Mistral-tiny}}\rightarrow$O$_{\text{Mistral-tiny}}\rightarrow$A$_{\text{Mistral-tiny}}$ &39.0 &47.5 &44.6 &72.0\\
  & Re$_{CoT}\rightarrow$ A$_{\text{Mistral-tiny}}\rightarrow$O$_{\text{Mixtral-8x22B}}\rightarrow$A$_{\text{Mistral-tiny}}$ &39.0 &48.2 &44.9 &\textbf{76.2}\\  
  & Re$_{CoT}\rightarrow$A$_{\text{Mistral-7B}}\rightarrow$O$_{\text{Mistral-7B}}\rightarrow$A$_{\text{Mistral-7B}}$ &40.0 &49.0 &45.5&75.3\\
  & Re$_{CoT}\rightarrow$A$_{\text{Mistral-7B}}\rightarrow$O$_{\text{Mixtral-8x22B}}\rightarrow$A$_{\text{Mistral-7B}}$ &\textbf{41.0} &\textbf{52.2} &\textbf{49.1} &72.0\\
  \hline
  \multirow{4}{*}{\rotatebox{90}{Closed}}   
  & Re$_{CaR}\rightarrow$A$_{\text{Gemini-Flash-8B}}\rightarrow$O$_{\text{Gemini-Flash-8B}}\rightarrow$A$_{\text{Gemini-Flash-8B}}$ &62.0 &67.6 &66.5& \textbf{78.5}\\
  & Re$_{CaR}\rightarrow$A$_{\text{Gemini-Flash-8B}}\rightarrow$O$_{\text{Gemini-Pro}}$\footnotemark[1]$\rightarrow$A$_{\text{Gemini-Flash-8B}}$ &\textbf{64.0} &\textbf{69.5} &\textbf{69.0} &74.0\\  
  & Re$_{CoT}\rightarrow$A$_{\text{Ministral-8B}}\rightarrow$O$_{\text{Ministral-8B}}\rightarrow$A$_{\text{Ministral-8B}}$ &59.0 &65.6 &64.8 &71.6\\
  & Re$_{CoT}\rightarrow$A$_{\text{Ministral-8B}}\rightarrow$O$_{\text{Mistral-large}}$\footnotemark[2]$\rightarrow$A$_{\text{Ministral-8B}}$ &60.0 &64.9 &64.2 &68.5\\
  \hline
  & Re$_{CoT}\rightarrow$A$_{\text{Mistral-tiny}}\rightarrow$O$_{\text{Gemini-Flash-8B}}\rightarrow$A$_{\text{Mistral-tiny}}$ & 40.0 &48.2& 45.4 &71.8\\
  \hline
  \end{tabular}}
\end{table}

\footnotetext[1]{Gemini version: 1.5}
\footnotetext[2]{Mistral version: mistral-large-2411}

When the same LLM is used for both the actor and optimizer, combining the actor and optimizer may lead to marginal improvements, making it more effective to use the actor alone. Additionally, even the more robust models sometimes have difficulty answering questions. They may indicate a lack of context or repeat the actor's response without providing additional information. In these situations, using an external tool might be helpful.

\subsubsection{Title generation: Impact of multimodal inputs}
The following experiments focused on multimodal inputs (i.e., text and image(s)). We used multimodal models to evaluate the actor in three different scenarios: (i) using textual input only (T), (ii) using multimodal input (MM), and (iii) a stepwise approach that uses textual followed by multimodal (T $\rightarrow$ MM) input. The actor uses zero-shot CoT and self-reflection as its reasoning mechanism, as this combination was the best-performed in our previous findings (~\autoref{table-MuSA-ablation-results-reason}). All experiments use the Gemini-1.5-Flash-8B model as the actor. The results are displayed in~\autoref{table-MuSA-ablation-results-MM}. 

\begin{table}[H] 
  \caption{Pass@1 100 EM, B4, RL on WikiWeb2M \citep{burns_suite_2023}. EM, B4, RL, F1, P, R: denote Exact Match, BLEU-4, ROUGE-L, F1-score, Precision, and Recall, respectively. T and MM denote textual and multimodal inputs, respectively. }
  \vspace{-5pt}
  \label{table-MuSA-ablation-results-MM}
  {\begin{tabular}{l r r r r r} 
  \hline 
  \textbf{Method} & \textbf{EM} & \textbf{B4} & \textbf{RL F1} & \textbf{RL P}& \textbf{RL R}\\
  \hline   
  A$_{T}$ &9&26.4&49.8&45.0&60.4\\
  A$_{MM}$ & \textbf{26} & \textbf{39.9}&\textbf{60.4}&\textbf{58.6}&\textbf{68.2}\\
  A$_{T} \rightarrow MM$ &\textbf{26}&\textbf{39.9}&59.5&57.9&67.2\\
 \hline
\end{tabular}}
\end{table}

Using multimodal information significantly enhances system performance in the task of title generation.

\subsubsection{Title generation: Impact of actor and optimizer roles for multimodal inputs}

We validate the performance of the actor and optimizer roles using two settings: (i) the actor and optimizer use the same multimodal LLM, and (ii) the optimizer uses a stronger multimodal LLM. The results are displayed in~\autoref{table-MuSA-ablation-results-MM-roles}.

\begin{table}[H] 
  \caption{Pass@1 100 EM, B4, RL on WikiWeb2M \citep{burns_suite_2023}.  EM, B4, RL, F1, P, R: denote Exact Match, BLEU-4, ROUGE-L, F1-score, Precision, and Recall, respectively. GF-8B and PX-12B denote Gemini-1.5-Flash-8B and Pixtral-12B-2409 models, respectively. } 
  \vspace{-5pt}
  \label{table-MuSA-ablation-results-MM-roles}
  \resizebox{\columnwidth}{!}
  {\begin{tabular}{l r r r r r} 
  \hline
  \textbf{Method} & \textbf{EM} & \textbf{B4} & \textbf{RL F1} & \textbf{RL P}& \textbf{RL R}\\
  \hline   
  Re$_{CaR}\rightarrow$ A$_{GF-8B}\rightarrow$O$_{GF-8B}\rightarrow$A$_{GF-8B}$ &27&35.2&51.3&47.2&63.2\\
  Re$_{CaR}\rightarrow$ A$_{GF-8B}\rightarrow$O$_{PX-12B}\rightarrow$A$_{GF-8B}$ &\textbf{32}& \textbf{41.6}&\textbf{61.3}&\textbf{59.0}&\textbf{71.9}\\
\hline
\end{tabular}}
\end{table}

In line with our previous findings (textual-only inputs) in~\autoref{table-MuSA-ablation-results-actor-eval}, when the same LLM is used for both the actor and optimizer, combining them may result in marginal improvements. Therefore, it may be more effective to use the actor alone. When the optimizer uses a more powerful multimodal LLM, the system benefits more.

%% file: 5.Discussion,v1.tex
\section{Discussion}

\textbf{MuSA. }Artificial intelligence (AI) aims at the design of systems that can
perform tasks that require human intelligence, and specialized AI systems are often developed with a specific task(s) in mind assigned to an agent(s). This work studies the assistant-user scenario, where a task is initially given. MuSA will conceptualize the task into a specific plan that consists of action(s) and complete it autonomously through conversations with other LLM-based units. Often, humans can have this initial idea of a task but may not possess the skills to accomplish it. An LLM-based agent specialized in this task can effectively complete it. In a multi-task environment, several instances of MuSA can be used to complete tasks independently. Though general-purpose LLMs are powerful, many real-world applications require only specific abilities and domain knowledge, and this modular approach offers several advantages over traditional monolithic single-agent systems. MuSA is designed to be modular and extensible, consisting of several independent LLM-based units (reason, plan, optimize, criticize, refine, and act) that can be easily integrated into the system and serve different purposes.

\noindent\textbf{Reason. } However, a significant concern when applying LLMs in critical domains is their tendency to hallucinate, generating factually incorrect or inconsistent responses while attempting to fulfill user requests. Step-by-step thinking is one of the most effective ways to improve performance and make the reasoning process more transparent. Adopting advanced reasoning prompting schemes, such as CoT and reflection, allows the actor/planner to reason and more efficiently solve a task. Our experiments have shown that the performance of each reasoning mechanism and their combination with the system seem to be model-specific. 

\noindent\textbf{Plan. } Planning in complex multimodal environments presents significant challenges for LLM-based agents. Many studies have proposed integrating memory modules to enhance the agents' robustness and adaptability. Long-term memory becomes crucial in these complex settings, while short-term memory is typically sufficient for agents managing tasks in simpler environments. Our work uses a short-term memory mechanism and has proven sufficient for its needs. 

\noindent\textbf{Optimize.} In a previous theoretical study \citep{cui_theoretical_2024}, the authors found that LLMs exhibit better error correction capacity and more accurate predictions when reasoning from previous steps is incorporated. Thus, combining the optimization unit with the planning/acting and reasoning processes has proven beneficial for our system, enhancing the agent's ability to learn from their mistakes and solve tasks. Furthermore, this design choice is helpful because one can use weaker models for the act or plan process, and optimization can be achieved using open-sourced models at a smaller cost, for example, rather than relying on stronger and more expensive models.

\noindent\textbf{Limitations. } The social content analysis tasks are numerous, and our system is initially designed to work with the four tasks listed in~\autoref{table-MuSA-actions}, which we have analyzed and presented in this study. We acknowledge this limitation and have included the development and analysis of additional social-content analysis actions in our plans. 

%% file: 6.Conclusion,v1.tex
\section{Conclusion}

This study introduces MuSA, an LLM-based multimodal agent that processes and analyzes online social content. Its computing engine integrates reason, plan, optimize, criticize, refine, and act processes which help MuSA to self-improve in complex scenarios. MuSA can plan based on social comprehension tasks, allowing it to execute more intricate tasks. Employing LLM agents can significantly assist human analysts by streamlining social content analysis-related tasks, eliminating the need for manual processes to discover and summarize social media discussions. Moreover, MuSA can potentially enhance and advance the quality of information in computational social listening systems.